%% file: root.tex
\documentclass[letterpaper, 10 pt, conference]{ieeeconf}  % Comment this line out if you need a4paper

\IEEEoverridecommandlockouts                              % This command is only needed if 
\usepackage{cite}

\title{\LARGE \bf
Distributed Client-Server Optimization for SLAM with Limited On-Device Resources
}

\author{Yetong Zhang$^{1,2}$, Ming Hsiao$^{1}$, Yipu Zhao$^{1}$, Jing Dong$^{1}$, and Jakob J. Engel$^{1}$% <-this % stops a space
\thanks{$^{1}$Facebook Reality Labs Research, Redmond, USA
        {\tt\small \{zhangyetong, mhsiao, yipuz, jingdong, jakob.engel\}@fb.com}}%
\thanks{$^{2}$College of Computing, Georgia Institute of Technology, Atlanta, USA
        {\tt\small yetong@gatech.edu}}%

}

\usepackage[dvipsnames]{xcolor}
\usepackage{graphicx}
\usepackage[ruled,vlined]{algorithm2e}
\usepackage{amsmath}
\DeclareMathOperator*{\argmax}{arg\,max}

\usepackage{tikz}

\begin{document}

\newcommand{\vn}{\Theta_n}
\newcommand{\vh}{\Theta_h}
\newcommand{\vs}{\Theta_s}
\newcommand{\vlc}{\Theta_{lc}}
\newcommand{\vho}{\Theta_{ho}}
\newcommand{\zn}{Z_n}
\newcommand{\zh}{Z_h}
\newcommand{\zs}{Z_s}
\newcommand{\zlc}{Z_{lc}}
\newcommand{\zns}{Z_{ns}}

\maketitle
\thispagestyle{empty}
\pagestyle{empty}

%%%%%%%%%%%%%%%%%%%%%%%%%%%%%%%%%%%%%%%%%%%%%%%%%%%%%%%%%%%%%%%%%%%%%%%%%%%%%%%%
\begin{abstract} 
%(The current abstract starts with too many background. Reader should be able to know what you do in this work in the first 1-2 sentences.)
Simultaneous localization and mapping (SLAM) is a crucial functionality for exploration robots and virtual/augmented reality (VR/AR) devices. However, some of such devices with limited resources cannot afford the computational or memory cost to run full SLAM algorithms. We propose a general client-server SLAM optimization framework that achieves accurate real-time state estimation on the device with low requirements of on-board resources. The resource-limited device (the client) only works on a small part of the map, and the rest of the map is processed by the server. By sending the summarized information of the rest of map to the client, the on-device state estimation is more accurate. Further improvement of accuracy is achieved in the presence of on-device early loop closures, which enables reloading useful variables from the server to the client. Experimental results from both synthetic and real-world datasets demonstrate that the proposed optimization framework achieves accurate estimation in real-time with limited computation and memory budget of the device.
% With the recent development in 5G network, the accessibility to fast and reliable communication has greatly improved, which motivates the application of client-server framework in simultaneous localization and mapping (SLAM) problems. The core challenge in such applications lies in the communication strategies, e.g. how the server can effectively send the information that is most helpful for the client, and how the client can request the information it needs most from the server. To this end, we make a systematic study on client-server SLAM framework, and propose strategies that use summarized measurements to send information from server to device \textcolor{red}{(few words that summarize our core contribution)}. We evaluate our methods on both synthetic and real-world datasets, and show our methods improve the state-of-the-art performance in achieving better accuracy on the device in real-time operations.
\end{abstract}

%%%%%%%%%%%%%%%%%%%%%%%%%%%%%%%%%%%%%%%%%%%%%%%%%%%%%%%%%%%%%%%%%%%%%%%%%%%%%%%%
\section{Introduction}
% \textcolor{gray}{Common SLAM algorithms do not apply to devices with limited resources.}
In the past decade, the applications of SLAM on light-weight devices have increased significantly, which include AR/VR headsets \cite{AR-SLAM}, small robots \cite{drone-SLAM} and smart phones \cite{slam_phone}. For such devices, the power consumption poses a strict bottleneck on the running time of general SLAM algorithms, which are usually expensive in computation and memory consumption. For future applications that require further scaling down of device size and power limit, such as coin-size drones \cite{drones_review,piccoli2017piccolissimo,giernacki2017crazyflie} and AR glasses, the applicability of conventional SLAM is greatly limited. For example, a state-of-the-art SLAM system, ORB-SLAM \cite{orb-slam}, normally runs on a multi-core CPU, and occupies memory up to 200MB. Meanwhile, a light-weight device such as a commercially available nano drone operates on an embedded system with less than 1MB of RAM \cite{giernacki2017crazyflie}. The gap between the on-device resource availability and the state-of-the-art SLAM requirement is huge.

 \begin{figure}[h!]
   \centering
   \includegraphics[width=0.4\textwidth]{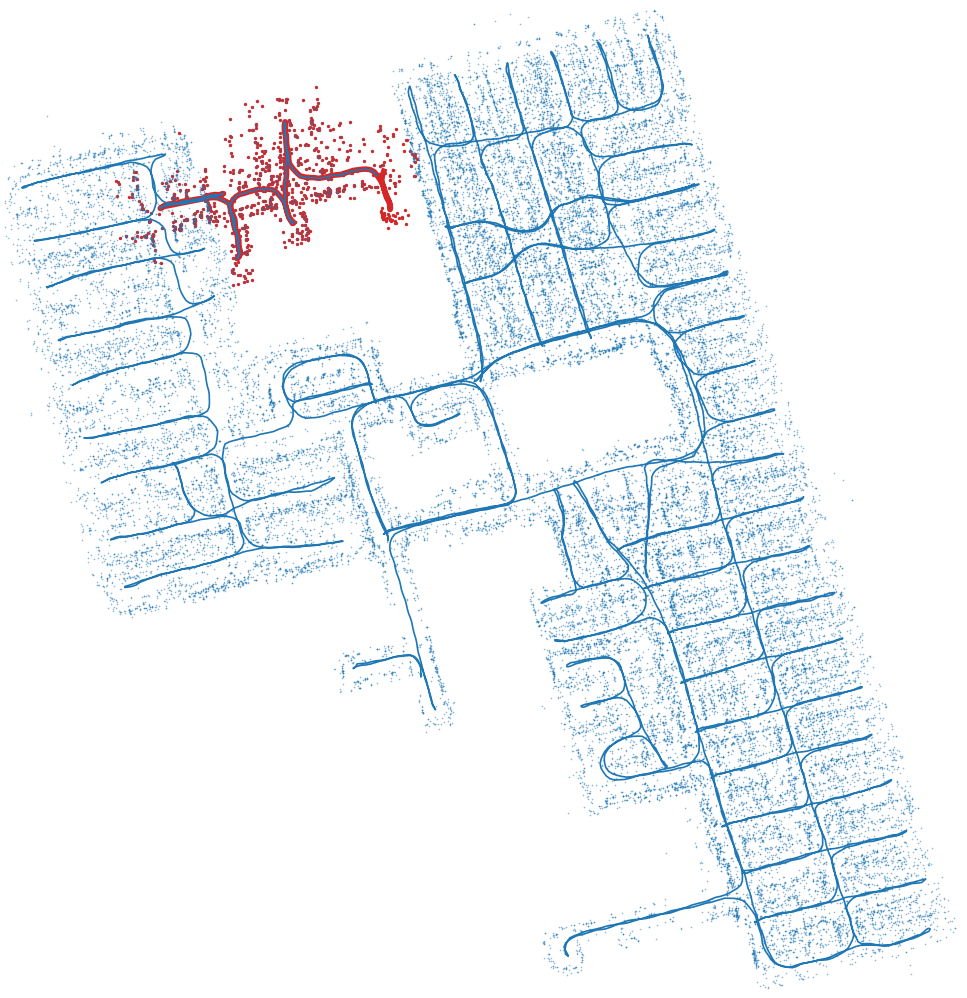}
 %   \resizebox{0.8\linewidth}{!}{\input{graphs/overall_fg.tikz}}
   \caption{An example of our proposed method running on a large indoor visual-inertial SLAM dataset. The entire area is around 50m$\times$70m. The blue trajectory and map points are maintained on the server side, and the red ones are maintained on the device side. More details about this dataset are discussed in Section \ref{sec:vins_dataset}.}
   \label{fig:exp}
 \vspace{-0.2in}
 \end{figure}

% \textcolor{red}{if we say in this way, it sounds like our algorithm is limited to applications of such small devices. If you are worried about this, we can add ``Even though this algorithm is developed for XXX, it also works on XXX without limitations" in the discussion subsection}
To achieve low computation and memory consumption, a common solution is to adopt sliding window or filtering approaches in SLAM. Such solutions only maintain a short memory of historic information, while dropping or marginalizing out-of-window information continuously. Due to the early fixation of linearization points and the accumulating nature of linearization error, the estimation accuracy of such solutions is limited \cite{sqrt-sam}. 

% However, most of the SLAM algorithms suffer from the trade-off between accuracy and \textcolor{gray}{efficiency (find a better word)} (this topic sentence/statement is a bit too complicated. The core idea is simply "SLAM is expensive"). 
% However, most of these devices lack the memory size and computation power to run full-map SLAM algorithm (what's the name of this algorithm?). To achieve accurate estimates, large-scale loop closures (should we directly use the term "loop closure"? or add some description) need to be considered, while detecting loop closures requires storing the history states, and running full-map optimization is computationally heavy. 

% \textcolor{gray}{Some existing works in client-server SLAM and their drawbacks}
In pursuit of a resource-efficient approach that does not suffer from aggregated linearization errors, several client-server SLAM frameworks \cite{hierarchical_slam} \cite{MOARSLAM} \cite{CCM_SLAM} have been proposed. In each of these works, a powerful server operates on all variables and measurements in the history (i.e., the full graph), and the resource-limited client works on a small fraction of variables and measurements (i.e., a local graph). Therefore, the device can achieve both computation and memory efficiency due to its small problem size, while it benefits from the more accurate estimation provided by the server.
Nevertheless, we argue that there are two limitations within the existing client-server SLAM frameworks.
Firstly, the device does not have any uncertainty information of the measurements outside its local graph, which loses the chances of improving the estimation results on the device with those information. Secondly, if an incoming measurement is associated with variables outside the local graph of the device (e.g. loop closure to an early estimation), the device cannot acquire such measurement until the server finishes optimization with the new variables. Therefore, the chance for the device to use the loop closure measurements to improve its estimation result is delayed.

To solve resolve the aforementioned drawbacks, we propose a client-server SLAM optimization framework with the following improvements from other existing works. Firstly, the server regularly sends summarized information of the entire history including the uncertainties to the device, which enhances the local graph that the on-device SLAM is working on. Secondly, the server is able to send back historic variables that are associated with the new variables on the device. As a result, the device can incorporate the measurements between the new and historic variables at an earlier time and perform an early loop closure thereafter. Thirdly, information sparsification is applied to the summarized information to bound the communication load. We evaluate our client-server SLAM optimization framework with both synthetic and real world datasets. The results show that our approach improves the real-time estimation results without increasing the computation load on the device.

%The contributions of this work include:
%\begin{itemize}
    %\item proposing a general distributed client-server optimization framework that incorporates summarized measurements to improve the state estimation accuracy on the device. 
    %\item applying information sparsification to client-server communication to avoid information congestion
    %\item extending the proposed algorithm to incorporate large loop closures at an earlier stage in the device \textcolor{blue}{Maybe order the contributions in the same way as they are introduced in the paper?}
    % \item An open source repository for public usage (link: XXXX) 
%\end{itemize}

\section{Related Work}
Client-server SLAM frameworks have been studied in several previous works.
% People have used client-server framework in SLAM problems to enable collaboration among multiple robots, and offload heavy computation from the device \textcolor{blue}{(In this paragraph, are we discussing "client-server SLAM", "multi-robot SLAM", or some fundamental method that they both share? Please be more clear)}. 
In \cite{client-server-slam-1}, a group of robots is coordinated by a server to explore unknown environment. Each robot performs localization on its own using particle filter, while the server collects the maps from each robot and proposes points of interest to guide the robots. In \cite{hierarchical_slam}, a graph-based approach was introduced to client-server SLAM framework. The server-side estimations are globally optimized, and shared with the client to rectify the client-side estimations. %The pose ground truth is estimated by a full visual-inertial bundle adjustment. 
In \cite{MOARSLAM} and \cite{CCM_SLAM}, the server acts as a central hub of all the information from the clients, enabling the clients to load adjacent maps to improve their measurements. 
In \cite{SLAMinDB}, the robot offload heavy computation to the server, and use the latest available pose from the server to reset its estimations.
However, none of these approaches incorporate the full information including uncertainties outside the local map in the device; nor did they make any systematic study on the communication timeline. 

One inspiration of this work comes from the existing study on decentralized multi-robot SLAM problems \cite{DDF_SAM,multi_robot_condensed_meas}, where the information were summarized and shared between robots in the form of condensed measurements. When one robot is sending information to another robot, it identifies the shared variables in their own graphs, and marginalizes out all the rest of the variables to generate marginal factors. If a server can use such methods to summarize the information outside the local graph of the device and provide it to the device, it is possible to improve the on-device estimations.

% \textcolor{blue}{(before jumping into the details of this work, add a line ahead to connect it to your work better. E.g.: ``Similar to the idea of C-S SLAM but operating on the same machine for speed(?), ...")} 
Another inspiration of this work comes from \cite{concurrent}. Even though it operates on a single machine, it is similar to a client-server approach because the inference problem is split into a filter component with lower latency but lower accuracy, and a smoother component with higher accuracy but higher latency. The full joint density is factorized in a way such that the two components of the inference problem are joined by a group of \emph{separator variables}, which is adopted in our client-server framework as well.

Information sparsification methods for SLAM problems have been well studied, and they can be applied in our framework to effectively bound the communication load. Two popular categories of information sparsification algorithms are generic linear constraints (GLC)\cite{carlevaris2013long,global_priors} and nonlinear factor recovery (NFR)\cite{mazuran2016nonlinear}. GLC approaches uses a set of linear factors to approximate the joint probability of the original factor, while NFR uses the nonlinear factors chosen by the user. In \cite{global_priors}, an efficient sparsification technique is proposed by ignoring the covariance between variables.
% Even though GLC is used in this work because of its generality, NFR should also be compatible with our framework as long as the type of nonlinear factors are determined based on the applications.

\section {Problem Formulation}
Before introducing the proposed methods, we first give a short introduction on 1) modeling a SLAM problem with a factor graph, and 2) how factor graphs are used in a client-server optimization setting as background.

\subsection{Factor Graph for SLAM Inference}

Factor graph~\cite{factor-graph} is a type of bipartite graph that consists of two types of nodes: variable nodes and factor nodes.
Each variable node represents an unknown variable $\theta \in \Theta$, where $\Theta$ is the set of all unknown variables; each factor node represents the potential $P(Z_i|\Theta_i)$ induced by a measurement $Z_i$, where $\Theta_i$ represents all variables associated to the measurement.
The entire factor graph represents a full joint probability $P(\Theta|Z)$ as Eq.~\ref{eqn:probability}.
As an inference problem, our goal is to compute the probability distribution of variables $\Theta$ provided with all the sensor measurements $Z$, and the maximum a posterior (MAP) estimate is acquired by Eq.~\ref{eqn:map}.
\begin{align}
    &P(\Theta|Z) \propto  \prod\limits_i P(Z_i|\Theta_i) \label{eqn:probability}\\
    & \Theta^* = \argmax\limits_{\Theta} \prod\limits_i P(Z_i|\Theta_i) \label{eqn:map}
\end{align}

\subsection{Client-Server Optimization}

\begin{figure}[!tb]
  \centering
  \resizebox{0.8\linewidth}{!}{\input{graphs/overall_fg.tikz}}
  \caption{
%   \textcolor{blue}{As the first figure of this paper, it can be a little misleading since your algorithm never constructs/optimizes this FG. Maybe move Fig.~4 to the 1st page and also refer to it briefly in the Intro, or combine Fig.~1 and Fig.~4 to illustrate how you decouple the original problem into the two parts?}
The factor graph representation of a toy client-server SLAM optimization problem. The server optimizes on the factor graph within the blue region, and the device optimizes on the factor graph within the red region. The separator variables $\vs$ and separator factors $\zs$ are marked in magenta, and they coexist in both the server and device. The factor marked in black is identified as a loop closure factor $\zlc$, because it associate with a new variable $\vn$ and a historic variable $\vh$.}
  \label{fig:overall-fg}
  \vspace{-0.1in}
\end{figure}
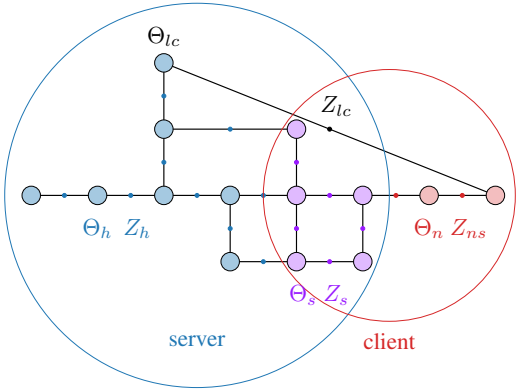

We formulate a client-server SLAM problem as a factor graph inference problem, in which the client and server have access to different parts of the factor graph (see Fig.~\ref{fig:overall-fg}).
At any given time, we define the variables and measurements that only exist in the server as \emph{historic variables} $\vh$ and \emph{historic measurements} $\zh$, the variables and measurements that exist in both the server and the device as \emph{separator variables} $\vs$ and \emph{separator measurements} $\zs$, the variables that only exist in the device as \emph{new variables} $\vn$. The measurements that associate with new variables are defined as \emph{new measurements} $\zn$. We further distinguish the new measurements that associate with some historic variables as \emph{loop closure measurements} $\zlc$, and define such historic variables as \emph{loop closure variables} $\vlc$. We use $\zns$, $\vho$ to represent the rest of new measurements and the rest of historic variables, respectively.

%Figure \ref{fig:overall-fg} shows the factor graph representation of the optimization problem in client-server framework. The server operates on the global scale by storing and optimizing over all the variables and factors in the history. The device operates on a  local  scale  by  only  storing  and  optimizing  a  small fraction of the variables and factors. \textcolor{blue}{Notice that due to the  transmission time from the device/client to the server and the longer optimization time on the server, some of the most recent variables and states in the device might not be in the server yet.} %The server adds new factors after a delay compared to the device, because the server takes longer time to run optimization on a larger factor graph, and it takes time to transmit the new measurements from the device to the server. Therefore, the factor graph in the device contains the most recent states and measurements which haven't been added to the factor graph on the server yet. 

Ideally the best possible estimation can be acquired by solving the full SLAM problem, which corresponds to the entire factor graph in Fig.~\ref{fig:overall-fg}.
\begin{align}
    \Theta^* = &\argmax\limits_{\Theta}
     P(\zh|\vh, \vs) P(\zs|\vs) \nonumber\\
    &P(\zlc|\vn, \vlc) P(\zns|\vn, \vs)
    \label{eqn:best_estimate}
\end{align}

However, such solutions cannot be acquired in existing client-server frameworks. Instead, the on-device optimization problem is performed only on the part of the factor graph that exists in the device:
\begin{align}
    \{\vn^*, \vs^*\} = 
    \argmax_{\vn, \vs}
    P(\zs|\vs) P(\zns|\vn, \vs)
    \label{eqn:existing_work}
\end{align}

Note that the problem in Eq.~\ref{eqn:existing_work} differs from the problem in Eq.~\ref{eqn:best_estimate} by missing two terms: $P(\zh|\vh, \vs)$, the full information of the factors and variables that only exist in the server, and $P(\zlc|\vn, \vlc)$, the loop closure measurements.
Both types of the missing information are helpful in the estimation of the new states $\vn$.

The goal of our framework is to provide the real-time state estimation as close as possible to the full SLAM solution in Eq.~\ref{eqn:best_estimate} with the limited computation resources on the device. As a result, we will take both $P(\zh|\vh, \vs)$ and $P(\zlc|\vn, \vlc)$ into account in the proposed framework.

% \begin{figure}[h!]
% \centering
% \resizebox{8cm}{!}{\input{graphs/overall_fg.tikz}}
% \caption{tikz test}\label{fig:fg_serial_four}
% \end{figure}

\section {Method}

\subsection{System Architecture}
\begin{figure}[!tb]
  \centering
  \includegraphics[scale=0.3]{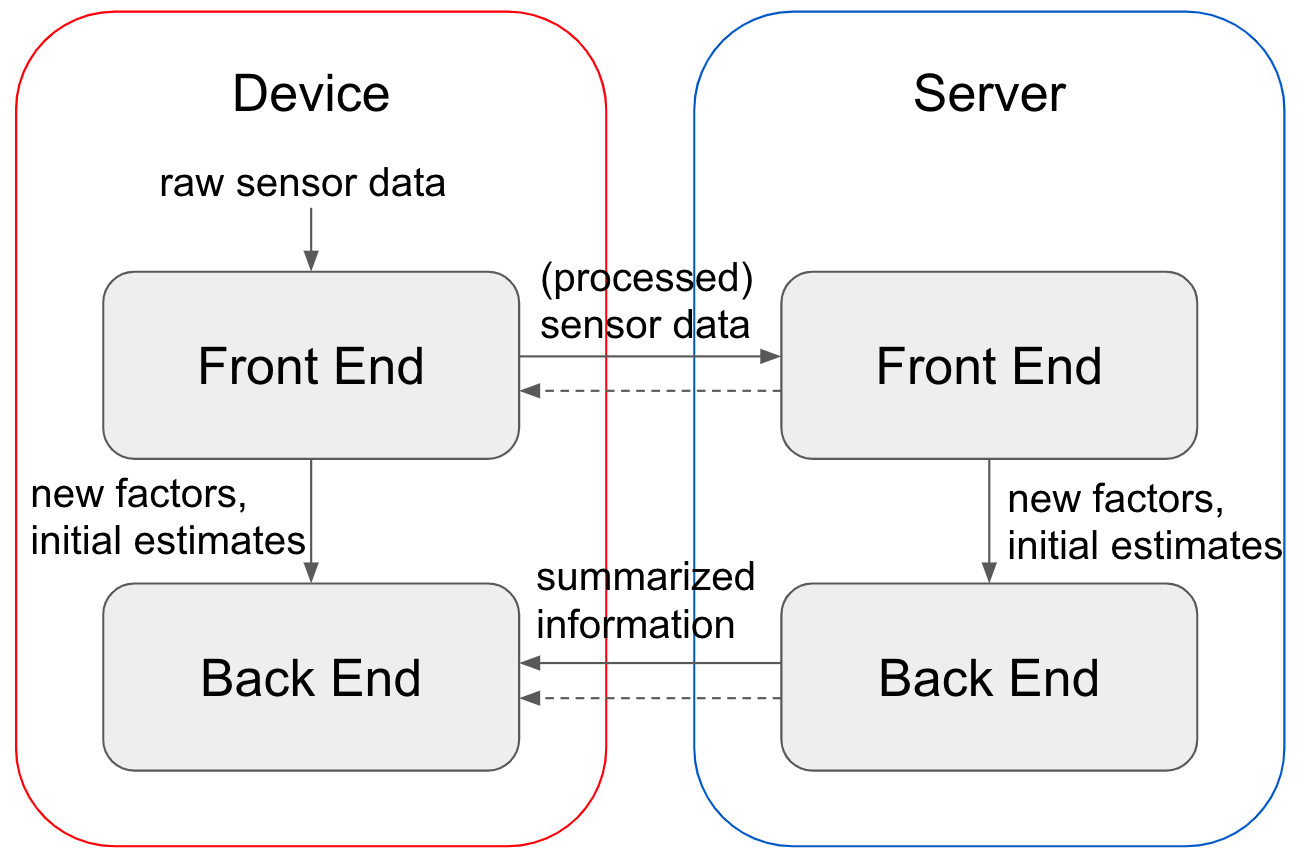}
  \caption{Overall system architecture: raw data is acquired from the sensors on the device, and is passed to the two front-ends that are on the device and the server respectively. Both front-ends process sensor data to generate new factors and initial estimates of new variables for the back end. The back-end of the server regularly sends summarized information to the back-end of the device. The dashed arrows represent the server sending back loop-closure information to the device (Section \ref{sec:early-lc}), where the back-end of the server sends the prior on loop closure variables and the front-end sends the loop closure factors and features associated with the loop closure variables. In the spatial implementation in Section \ref{sec:spatial-temporal}, the dashed arrows additionally represent the server reloading ``nearby" graph to the device.}
  \label{fig:system-structure}
  \vspace{-0.1in}
\end{figure}

In our framework, a client and a server run a SLAM algorithm together in a collaborative way. The goal of the client is to provide real-time estimates. The goal of the server is to provide accurate estimates, which can be used by the device to rectify its estimation results. The overall system structure is shown in Fig.~\ref{fig:system-structure}. 

Both the device and the server have a front-end and a back-end. 
Both front-ends operate in a similar way: they process the sensor data, create factors corresponding to the measurements derived from the sensor data, and provide initial estimates of the new variables associated with the measurements.
However, the factors and the timing of generating them are different in these two front-ends.
Since the client stores both the new and separator variables, the client front-end can add the new factors among them, $P(\zns|\vn,\vs)$, to the client factor graph immediately to provide fast real-time estimates.
However, since only the server stores all variables, the loop closure measurements $\zlc$ that associated with variables outside the scope of the device can only be discovered in the front end of the server. Therefore, the server front-end adds loop closure factors, $P(\zlc|\vn,\vlc)$, in addition to the same new factors added in the client at a later time to provide slower but more accurate estimates. 
%because only the server stores all the history variables, and the factors related to the variables outside the scope of the device can only be discovered by the front-end in the server.

Both back-ends maintain their own factor graphs that collect the factors generated by their corresponding front-ends, and compute the MAP estimates of the variables in their factor graphs. 
The server back-end keeps all the variables and factors, while old variables are removed from the device back-end to bound the memory size.

\subsection{Summarized Information} \label{sec:summarized_information}
In factor graphs, a common approach to preserve only part of the variables while keeping the rest information approximated is marginalization \cite{probabilistic-robotics}. By applying marginalization on a factor graph, the full joint probability is factorized into the form of $P(\Theta_r, \Theta_e) = P(\Theta_r)P(\Theta_e|\Theta_r)$, where $\Theta_e$ represents the variables to be marginalized out, and $\Theta_r$ represents the remaining variables.

\begin{figure}[h!]
  \centering
    \resizebox{0.9\linewidth}{!}{\input{graphs/marginalization.tikz}}
  \caption{The summarized information is generated by the server, and incorporated in the device. 1) The server marginalizes the history variables marked in blue, and generate marginal factors (marked in pink) on the separator variables. 2) The server sends the marginalization factor to the device. 3) The device remove old variables and factors marked in blue, and update the marginalization factor.}
  \label{fig:device-server-fg}
  \vspace{-0.1in}
\end{figure}
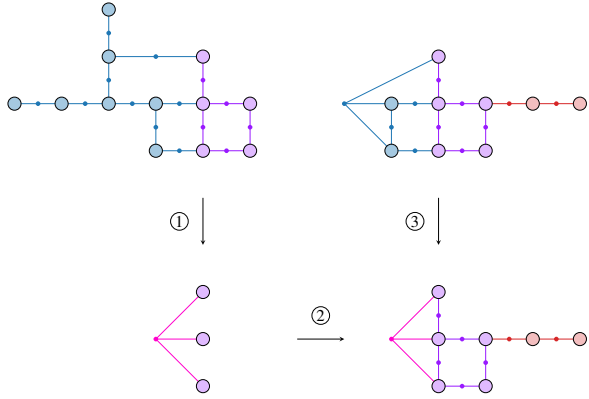

Fig.~\ref{fig:device-server-fg} illustrates the process of the server generating the summarized information, and the device incorporating the summarized information in its factor graph. To summarize the part of the factor graph that is unavailable in the device, we can apply Eq.~\ref{eqn:server_marginalization} to marginalize out the historic variables $\vh$ and get the remaining marginal term $P(\vs|\zh)$, which is correct up to the fixed linearization point.
The linearization point is expected to be the best possible estimate at each time step, since the server performs global optimization on all the accumulated measurements. Therefore, we can pass the marginalization factor $P(\vs|\zh)$ to the device as a theoretically accurate approximation of the rest of the graph.
\begin{align}
    P(\vs, \vh|\zh) = P(\vs|\zh)P(\vh|\vs, \zh)
    \label{eqn:server_marginalization}
\end{align}

In practice, it is desirable to use an inference algorithm on the server that is efficient in both incremental optimization and generating summarized information, as the execution time of the two processes is directly related to the delay for the server to send out the summarized information. 
In our implementation, we use the iSAM2 algorithm~\cite{isam2} for the server in our implementation, as it excels in incremental sparse nonlinear optimization by utilizing the Bayes tree structure. It is noted in \cite{DDF-SAM2} that the Bayes tree data structure also provides convenience in generating marginal factors.

% In practice, since the time for both the incremental optimization and the marginalization to generate the summarized information are directly related to the lag $t_d$ in the server, it is desired to have an efficient algorithm to reduce the delay. The iSAM2 algorithm \cite{isam2} excels in sparse nonlinear optimization by utilizing the Bayes tree structure. Following the variable ordering in \cite{isam2}, the newer variables are positioned nearer to the root of the Bayes tree. Therefore, only the top part of the tree are updated in most cases, which allows fast incremental optimization.

% The Bayes tree structure also improves efficiency in marginalization. Since every node represents a clique of variables, every non-root clique shares a set of frontal variables with its parent clique, and sends information to its parent clique, which contains the marginal factors on the frontal variables by marginalizing out the rest of variables in the clique. As shown in \cite{DDF-SAM2}, we can take advantage of these pre-computed inter-clique marginals to speed up the process of generating marginal factors.

\subsection{Early Loop Closure} \label{sec:early-lc}
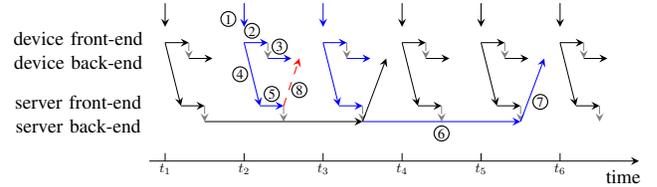
\begin{figure}[thpb]
  \centering
    \resizebox{1.0\linewidth}{!}{\input{graphs/communication.tikz}}
  \caption{A full cycle of updating estimates in our client-server system is marked in blue. 1) New sensor data is generated on the device at certain frequencies. 2) The device front-end processes the raw sensor data and generates factors $\zns$. 3) The device back-end performs a quick local optimization with the new measurements included. 4) At the same time, the device also sends the sensor data to the server. 5) The server front-end generates all factors including loop-closure factors ($\zns \cup \zlc)$. It will cache the factors until the server back-end is ready to add them to the factor graph. 6) The back-end of the server performs optimization on the entire factor graph with cached factors, and generate summarized information. 7) The server sends summarized information to the device. 8) The additional component for early loop closure method is marked in red. Upon detecting loop closure factors, the server sends loop closure information to the device.}
  \label{fig:communication-timeline}
  \vspace{-0.1in}
\end{figure}

Through inspecting the communication schedule of a standard approach described in Section \ref{sec:summarized_information} (Fig.~\ref{fig:communication-timeline}), we discovered that there is an unavoidable delay $t_d$ for the device to incorporate the loop-closure measurements $\zlc$ into optimization after the new variables $\vn$ are firstly observed in the sensor. Since the device do not store historic variables $\vh$, it can only get loop-closure information from the server. The total delay is composed of 4 parts: $t_d = t_s + t_f + t_b + t_r$, where $t_s$ is the time to send measurement information from device to server; $t_f$ is the operation time in the server to generate factors in the front-end; $t_b$ is the time to perform optimization and summarize information in the back-end; $t_r$ is the time to send summarized information from the server to the device. Normally, the back-end optimization on the server is the most time consuming part in $t_d$ due to the large factor graph size in the server. This operation is especially time consuming when large loop closure is encountered, because lots of variables are involved in the loop.

Therefore, we propose methods to reduce the delay so that the server can incorporate such measurements at an earlier time. We note in Fig.~\ref{fig:communication-timeline} that the loop closure factors are generated before the server starts optimization. Therefore, the server can send additional loop closure information to the device right after the loop closure factors are detected (see the red arrow in Fig.~\ref{fig:communication-timeline}), reducing the delay to $t_s+t_f+t_r$ by skipping the server back-end optimization time $t_b$.

To make use of the loop closure factors, the device needs to know the priors on the loop closure variables $\vlc$. 
As the server already passes the marginal on the separator variables $P(\vs|\zh)$, we can further apply a factorization to Eq.~\ref{eqn:server_marginalization} as
\begin{align}
    &P(\vs, \vh|\zh) \nonumber\\
    =&P(\vs|\zh)P(\vlc|\vs, \zh)P(\vho|\vs, \vlc, \zh) \label{eqn:lc_marginalization}
\end{align}
Therefore, the posterior probability $P(\vlc|\vs, \zh)$ can be sent to the device to incorporate prior information of loop closure variables $\vlc$, while maintaining consistency.
% Theoretically, the probabilistic exact approach to pass prior information from the server to the device is to compute the posterior $P(\vlc|\vs, \zh)$, which can be derived by applying a factorization on Eq.~\ref{eqn:best_estimate} as shown in Eq.~\ref{eqn:marginal-lc}. 

However, as the posterior $P(\vlc|\vs, \zh)$ can be costly to compute, we approximate it with $P(\vlc| \zh)$, which is easy to compute. Notice that this approximation ignores the covariance between $\vlc$ and $\vs$, which is similar to the information sparsification strategies used in \cite{global_priors}. Even though the approximation is not guaranteed to be probabilistic conservative, it will be overwritten by the theoretically correct summarized information from the server shortly after.

\section{Implementation Details}
\subsection{Spatial vs. Temporal} \label{sec:spatial-temporal}
The choice of separator variables can greatly affect the performance of the proposed algorithm. Since the front-end of the device can only discover the measurements associated with separator variables $\zns$, choosing better separator variables can enable the front-end to incorporate more measurements and generate more factors accordingly. Moreover, we would like to keep the separator variables that are more likely to have a larger update by the new measurements, so that we have a chance to relinearize the factors among them to provide better estimates.

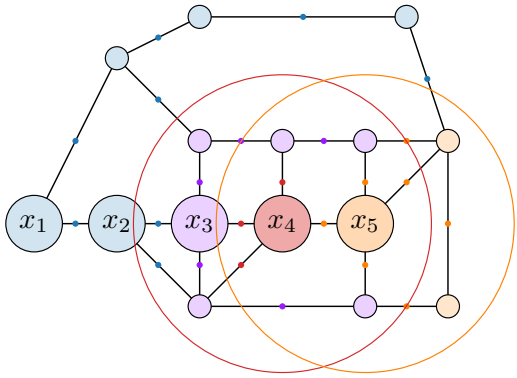
\begin{figure}[!tb]
  \centering
  \resizebox{0.8\linewidth}{!}{\input{graphs/spatial_fg.tikz}}
  \caption{An example demonstrating the spatial heuristic. Suppose the robot follows a trajectory $x_1$ - $x_5$ through a previously visited area. When the robot is at $x_4$, the adjacent region is marked by the red circle, and the separator variables are marked in magenta. When the robot moves to $x_5$, the variables and factors that become adjacent to the robot are marked in orange. Notice that such historic variables as well as the factors among them need to be reloaded to the device from the server.}
  \label{fig:spatial_fg}
  \vspace{-0.2in}
\end{figure}

There are two heuristics to select separator variables: the most recent ones (temporal), or the most adjacent ones (spatial). The implementation for temporal case is straightforward, since all variables are added in a temporal order. As for the spatial case, more work needs to be done. When the robot re-visits a previously observed region, some spatially adjacent variables may have been previously eliminated in the device, and becomes historic variables $\vh$ (see Fig.~\ref{fig:spatial_fg}). These nearby historic variables, as well as the measurements among them, need to be reloaded to the device. 

In general, the spatial heuristic can generate more factors in the device than the temporal heuristic, since the new measurements are commonly related to the adjacent variables. Therefore, some loop closure measurements in the temporal case can be discovered in the front end with spatial heuristic.

However, the spatial heuristic may suffer from drifting, and it is more sensitive to communication delay than the temporal heuristic. In the case of severe drifting, the estimation of current pose drifts from the actual pose. Therefore, the adjacent variables identified by the current estimates may not be adjacent in the real world. In the case of large communication delay, the adjacent region identified by the server may no longer be adjacent to the device by the time the device receives the information, as the device may have moved out of the region during the delay $t_d$.

%\textcolor{red}{include algorithms for the server/device here?}

% \textcolor{red}{Should I include the detailed algorithm here?} \textcolor{blue}{What alg?}

% \begin{algorithm}[h]
% \SetAlgoLined
% \SetKwInOut{Input}{input}
% \SetKwInOut{Output}{output}
% \Input {Values of separator variables, marginal factors, additional graph\\}
% % \Output {Variables marginalized out}
% Remove the current marginalization factors\;
% Remove factors in the current graph that only associate with separator variables\;
% Add the incoming marginal factors to the graph\;
% Perform optimization\;
% Further select old variables to marginalize out\;
%  \caption{Device update in spatial case}
%  \label{spatial-device}
% \end{algorithm}

\subsection{Information Sparsification}
In Sec. IV-B, since the marginal factor is usually dense, its size, as in the form of an information matrix, is proportional to $n^2$, with $n$ representing the number of variables it connects to. Therefore, when we send the marginal factors to the device, the message size may be too large, which increases the computation and memory costs in the device, and can potentially cause information congestion. 

% In order to reduce the message size, one way is to reduce the size of separator variables. Reducing the separator variable size may (some words missing?) performance, (NO ,) since loop closures can only be performed in a smaller scale in the device. (unclear)

Therefore, we adopt information sparsification methods on the marginal factors, which generates a set of small sized factors to approximate the probability density induced by the marginal factors. We desire the algorithm to be simple, because the time for sparsification is part of $t_b$ (in addition to the optimization and marginalization time), which contributes to the delay $t_d$. The algorithms that compare the KL-divergence for each pair of variables, such as Chow-Liu tree \cite{ChowLiu} is too computationally expensive for our application. As a result, we adopt the efficient Global Priors \cite{global_priors} sparsification approach in our implementation. We also show in the experiments that even assuming the same delay $t_d$, Global Priors generates similar results as the Chow-Liu tree sparsification methods in our applications.

\section{Experiments}
\subsection{Benchmarking setup}
We implement the proposed algorithm based on GTSAM~\cite{dellaert2012factor}, which is a general C++ factor graph optimization library with an iSAM2 implementation.
Both synthetic and real-world datasets are used to evaluate the performance of the proposed framework. The benchmarking is conducted in a simulation environment, where two threads are allocated to simulate server and client processes respectively. The server-client communication channel is simulated by passing message between the two threads. All evaluations are performed on a laptop computer.

We compare various configurations of our proposed method with a baseline, which simply uses the latest pose sent from the server to reset its estimations as in \cite{SLAMinDB}. The six configurations include temporal separators, spatial separators, temporal separators with sparsification (marked as + S), temporal separators with early loop closure (marked as + LC), and temporal/spatial separators with both sparsification and early loop closure (marked as + S + LC). The pseudo ground-truth we adopt to evaluate accuracy is the batch optimization results at each iteration, since they are the best possible estimates with \emph{all} the available information to date as in Eq.~\ref{eqn:best_estimate}. 

For all configurations, We show four major metrics for evaluation: average (per frame) translation error and rotation error for the real-time state estimation results on the device compared to pseudo ground-truth, average server processing time ($t_b$), and the average message size sent from server to device (amount of floating point numbers). Notice that we do not show the comparison of device side optimization time since the device in each experiment has very similar graph and variable size that result in very close time consumption. The additional loop closure factors have little influence on the graph size of the device because the number of loop closure variables $\vlc$ is much fewer than the number of separator variables $\vs$ in most cases, and $\vlc$ can be eliminated in advance in the device, since they only connect to global priors and loop closure factors.

%he real-time state estimation results generated from the device thread is compared against the result generated by an incremental optimization algorithm, which incorporates all the history measurements as in (\ref{eqn:best_estimate}), and is the best possible estimation the device can get. We also include the server optimization time that is directly related to the delay $t_d$, and the size of the summarized information.

\subsection{Synthetic 2D pose graph dataset}

The first dataset we use is the M3500 synthetic dataset introduced in~\cite{olson2006fast}, which is a 2D pose graph dataset consisting of 3500 poses and 5600 relative pose constraints. we assume the incoming measurements emerge at a constant rate of $20ms$, with 10 new states at each step. We additionally fix the communication lag to $10ms$, and the size of separator variables to 300 poses. 

From the benchmark results in Table~\ref{table:benchmark-posegraph}, we see that all configurations perform better than the baseline in terms of estimation accuracy. Since new states are most likely to associate with adjacent variables, the spatial configuration can incorporate more measurements and result in better estimation accuracy than the temporal configuration. Moreover, since the marginal factors in the spatial case are not as densely connected as in the temporal case, its message size is smaller. We further observe that information sparsification results in much smaller message sizes.

\begin{table}[!tb]
\caption{Comparison of client-server algorithms on 2D synthetic pose graph dataset}
\vspace{-5mm}
\label{table:benchmark-posegraph}
\begin{center}
\begin{tabular}{l|cccc}
\hline
Algorithm & \begin{tabular}{@{}c@{}}Translation\\error (m)\end{tabular}&\begin{tabular}{@{}c@{}}Rotation\\error (rad)\end{tabular}&\begin{tabular}{@{}c@{}}Server\\time (s)\end{tabular}&\begin{tabular}{@{}c@{}}Message\\({\#float})\end{tabular} \\
\hline
Baseline & 0.407 & 0.0343 & 0.086 & 900\\
Temporal & 0.357 & 0.0311 & 0.082 & 18725\\
Spatial &  0.229 & 0.0196 & 0.082 & 5226\\
Temporal + S & 0.331 & 0.0286 & 0.089 & 1336\\
Temporal + LC & 0.235 & 0.0218 & 0.083 & 18834\\
Temporal + S + LC &  0.192 & 0.0191 & 0.088 & 1451\\
Spatial + S + LC & 0.191 & 0.0192 & 0.081 & 3280 \\
\hline

\end{tabular}
\end{center}
\vspace{-0.4cm}
\end{table}

\begin{figure}[!tb]
  \centering
    \includegraphics[scale=0.17]{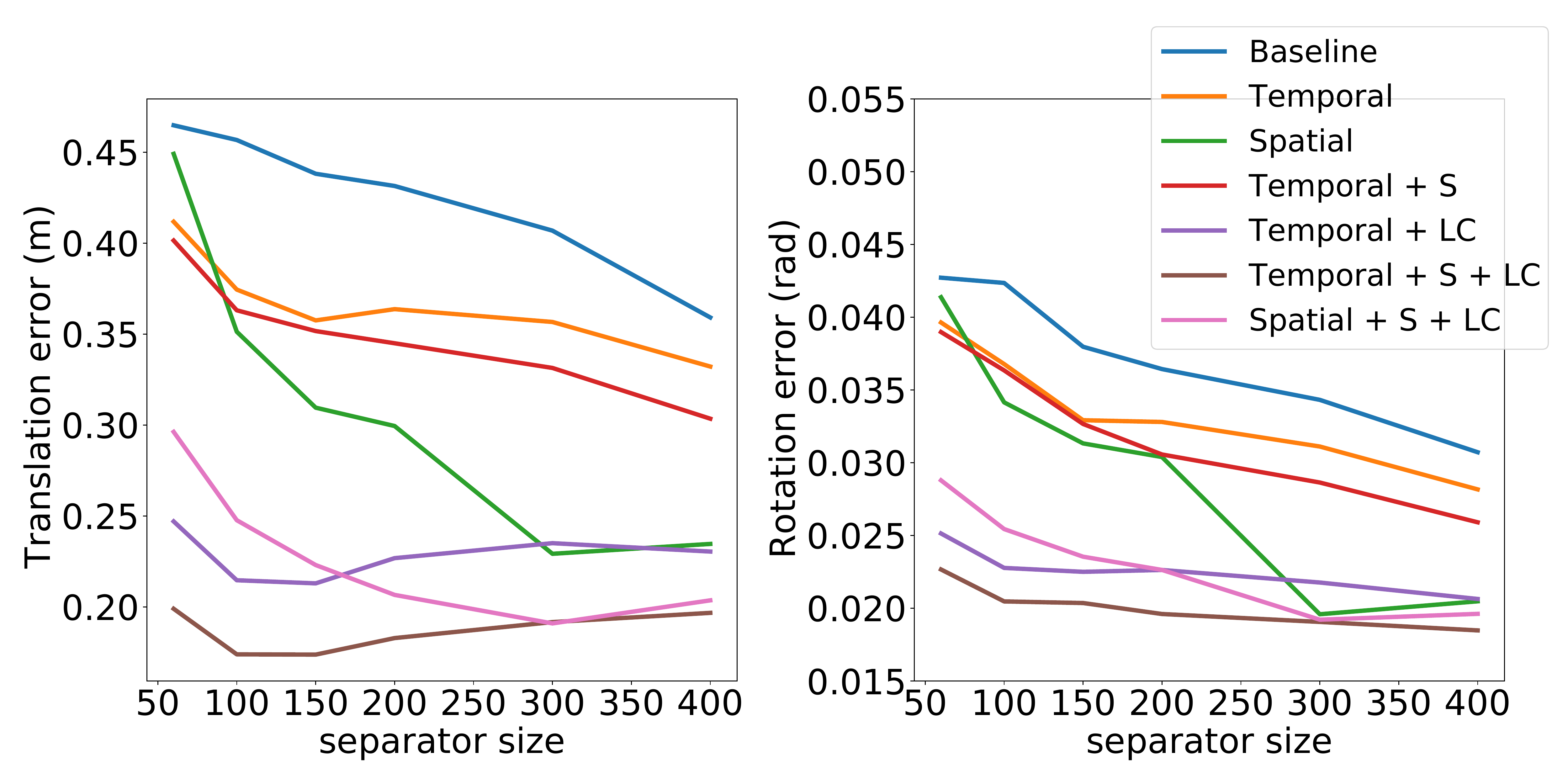}  
  \caption{Plot of the client estimation accuracy vs. the size of separator variables $|\vs|$ for M3500 dataset.}
  \label{fig:separator-size-vs-error}
  \vspace{-0.5cm}
\end{figure}

Furthermore, we study how the size of separator variables influences the estimation accuracy (see Fig.~\ref{fig:separator-size-vs-error}). Since larger separator variables size means more information is sent to the device, the estimation accuracy should be improved given larger separator variables size. The trend is clearly shown in the rotational and translational error plot. The performance with early loop closure helps a lot when separator size is small since it gains additional loop closure information for real-time state estimations. As the separator size increases, more loop closure measurements are likely to be incorporated in the device, and the effect of early loop closure is smaller. Therefore, the early loop closure strategy is most helpful when the graph size in the device is extremely limited.

It is interesting to discover that on this dataset, sparsification does not hurt the on-device estimation accuracy, and it even decreases the estimation error when early loop closure is applied, as ``Temporal+S+LC" consistently outperforms ``Temporal+LC" in Fig.~\ref{fig:separator-size-vs-error}.
One possible explanation is that the adopted sparsification method \cite{global_priors} uses the same approximation strategy as the one used to generate priors for loop closure variables $\vlc$, which simply ignores the covariance between variables. Since the approximation strategy may not be conservative, it is possible that the non-conservative approximations of the priors on the separator variables $\vs$ and loop closure variables $\vlc$ are "balanced", therefore resulting in better estimates.
% is know to generate over-confident sparsified factors, the early loop closure factors computed from it would constrain the variables in the device more than the marginalized factors on the separators do. However, when we apply \cite{global_priors} to both the marginals of the separators and the early loop closure factors, all the constraints are over-confident yet "balanced", therefore achieves better estimates.

It is worth noting that the message size is only computed for the back-end communication. The communication in front-ends depends on the specific implementation that addresses data association.
% \begin{figure}[h!]
%   \centering
%   \includegraphics[scale=0.45]{imgs/delay-error.png}
%   \caption{Influence of communication delay on client estimation accuracy }
%   \label{fig:delay-vs-error}
% \end{figure}

% \begin{figure}[h!]
%   \centering
%   \includegraphics[scale=0.45]{imgs/fig_compare1.png}
%   \caption{experiment result (comment: treat the figure as a place holder)}
%   \label{fig:exp-result}
% \end{figure}

\subsection{Real-world visual-inertial SLAM dataset} \label{sec:vins_dataset}

The second dataset we use to evaluate our approach is a visual-inertial SLAM dataset, which is collected internally at Facebook.
The sequence is collected by a custom rig, where multiple cameras and an IMU are mounted on a headset rigidly.
The size of the mapped area is about 50m by 70m, and the raw sensor stream sequence is about 30 minutes long. We get visual landmark information and pose-to-pose constraints from an in-house key-frame based front-end. In Fig.~\ref{fig:exp} we show the server and device map built by our algorithm (temporal separators configuration) close to the end of the sequence. The new measurements are sent to the server and the device at approximately 0.3s per 10 key frames, and we set the communication delay to 0.1s.

\begin{table}[!tb]
\caption{Comparison of client-server algorithms on real-world visual-inertial dataset}
\vspace{-5mm}
\label{table:benchmark-realworld}
\begin{center}
\begin{tabular}{l|cccc}
\hline
Algorithm & \begin{tabular}{@{}c@{}}Translation\\error (m)\end{tabular}&\begin{tabular}{@{}c@{}}Rotation\\error (rad)\end{tabular}&\begin{tabular}{@{}c@{}}Server\\time (s)\end{tabular}&\begin{tabular}{@{}c@{}}Message\\({\#float})\end{tabular} \\
\hline
Baseline & 0.165 & 0.0172 & 1.29 & 663\\
Temporal & 0.142 & 0.0157 & 1.40 & 64083\\
Spatial & -- & -- &--  &-- \\
Temporal + S & 0.169 & 0.0171 & 1.46 & 1820\\
Temporal + LC & 0.129 & 0.0139 & 1.38 & 65220\\
Temporal + S + LC & 0.129 & 0.0140 & 1.45 & 2237\\
Spatial + S + LC & -- & -- & -- & --\\
\hline

\end{tabular}
\end{center}
\vspace{-0.5cm}
\end{table}

% The benchmarked algorithm configurations are the same as in the pose graph experiment.
% : 
% fixed lag smoother as baseline; temporal separators + global priors sparsification setting; temporal separators + early loop closure setting; and temporal separators + global priors sparsification + early loop closure setting. 
Since we have much denser graph in this visual-inertial SLAM dataset than the previous pose graph dataset, the device can only keep a smaller number of separator variables, which is set as 200 in all configurations. 

From the results in Table~\ref{table:benchmark-realworld}, we can again observe that incorporating the marginal uncertainty of temporal separators outperforms the baseline in accuracy. Moreover, temporal with sparsification and early loop closure increases estimation accuracy over the baseline without severely increasing the message size. The advantages of early loop closure would not have been acquired properly if we do not have uncertainty information over the separator variables. 

However, both spatial cases fail due to communication delay. By the time the device receives the summarized information form the server, the current location of the device has moved out of the "adjacent region" marked by the separator variables. Therefore, the device fails to connect the new variables with the separator variables, and cannot provide valid estimates. In practice, the device can reject the summarized information from the server, and operate as a fixed-lag smoother in such circumstances.

%\subsection{Discussion}
%By utilizing the marginal information from the server, the device can incorporate the uncertainty information of the separator variables, thus generating better estimation results than 

%// TODO

% There exists a trade-off between fixed-lag accuracy (what is "fixed-lag accuracy"?) and communication lag. The spatial-case communication yields smaller estimation error with fixed communication lag, while it requires heavier server-side computation and a larger message size, which would in turn increase the communication lag and thus decreasing the client-side estimation accuracy. Similarly, increasing the size of separator variables improve the client-side accuracy, while increase the communication lag. Therefore, when the communication is stable and fast, algorithms that sacrifice the communication for better estimation accuracy is preferred, and vice versa.

% \begin{figure}[h]
% \centering
% \resizebox{8cm}{!}{\input{graphs/overall_fg.tikz}}
% \caption{tikz test}\label{fig:fg_serial_four}
% \end{figure}

%\section{FUTURE WORK}
%The current iSAM2 algorithm is not efficient in generalizing to multi-device settings, since all the information aggregated will be added to the root of the tree. Merging them at the same root will not only destruct the natural sparse structure of the problem, but also make it harder to summarize information for a specific device. The next step will be to devise a multi-root iSAM2 algorithm which enables update at multiple branches of the tree. \textcolor{blue}{(Maybe also talk about saving power?)}

\section {Conclusion}
In this paper, we propose a general client-server optimization framework to solve SLAM problems with limited on-device resources.
The summarized information is sent from the server to the device to inform the device of the uncertainty information outside of its local map.
By carefully inspecting the communication timeline, we propose methods to achieve on-device early loop closure.
We further perform a systematic study on the temporal and spatial heuristics in selecting separator variables, and the influence of applying sparsification on the summarized information.
Future works include applying the proposed framework in a real-world system, refining the framework to consider on-device power efficiency, and generalizing the server to work with multiple devices simultaneously.

\bibliographystyle{plain}
\bibliography{refs}

\end{document}

%% file: graphs/overall_fg.tikz
\begin{tikzpicture}
\definecolor {def_purple} {rgb} { 0.6,0.1, 1.0 };
\definecolor {def_blue} {rgb} { 0.12156862745098039,0.4666666666666667,0.7058823529411765 };
\definecolor {def_red} {rgb} { 0.8392156862745098,0.15294117647058825,0.1568627450980392 };
\coordinate (v1) at (0, 0);
\coordinate (v2) at (1, 0);
\coordinate (v3) at (2, 0);
\coordinate (v4) at (3, 0);
\coordinate (v5) at (4, 0);
\coordinate (v6) at (5, 0);
\coordinate (v7) at (6, 0);
\coordinate (v8) at (7, 0);
\coordinate (v9) at (3, -1);
\coordinate (v10) at (4, -1);
\coordinate (v11) at (5, -1);
\coordinate (v12) at (2, 1);
\coordinate (v13) at (4, 1);
\coordinate (v14) at (2, 2);
\coordinate (f1) at (0.5, 0.0);
\coordinate (f2) at (1.5, 0.0);
\coordinate (f3) at (2.5, 0.0);
\coordinate (f4) at (3.5, 0.0);
\coordinate (f5) at (4.5, 0.0);
\coordinate (f6) at (5.5, 0.0);
\coordinate (f7) at (6.5, 0.0);
\coordinate (f8) at (3.0, -0.5);
\coordinate (f9) at (4.0, -0.5);
\coordinate (f10) at (5.0, -0.5);
\coordinate (f11) at (3.5, -1.0);
\coordinate (f12) at (4.5, -1.0);
\coordinate (f13) at (2.0, 0.5);
\coordinate (f14) at (4.0, 0.5);
\coordinate (f15) at (2.0, 1.5);
\coordinate (f16) at (3.0, 1.0);
\coordinate (f17) at (4.5, 1.0);
\path[draw, line width=0.5pt] (v1) -- (f1);
\path[draw, line width=0.5pt] (v2) -- (f1);
\path[draw, line width=0.5pt] (v2) -- (f2);
\path[draw, line width=0.5pt] (v3) -- (f2);
\path[draw, line width=0.5pt] (v3) -- (f3);
\path[draw, line width=0.5pt] (v4) -- (f3);
\path[draw, line width=0.5pt] (v4) -- (f4);
\path[draw, line width=0.5pt] (v5) -- (f4);
\path[draw, line width=0.5pt] (v5) -- (f5);
\path[draw, line width=0.5pt] (v6) -- (f5);
\path[draw, line width=0.5pt] (v6) -- (f6);
\path[draw, line width=0.5pt] (v7) -- (f6);
\path[draw, line width=0.5pt] (v7) -- (f7);
\path[draw, line width=0.5pt] (v8) -- (f7);
\path[draw, line width=0.5pt] (v4) -- (f8);
\path[draw, line width=0.5pt] (v9) -- (f8);
\path[draw, line width=0.5pt] (v5) -- (f9);
\path[draw, line width=0.5pt] (v10) -- (f9);
\path[draw, line width=0.5pt] (v6) -- (f10);
\path[draw, line width=0.5pt] (v11) -- (f10);
\path[draw, line width=0.5pt] (v9) -- (f11);
\path[draw, line width=0.5pt] (v10) -- (f11);
\path[draw, line width=0.5pt] (v10) -- (f12);
\path[draw, line width=0.5pt] (v11) -- (f12);
\path[draw, line width=0.5pt] (v3) -- (f13);
\path[draw, line width=0.5pt] (v12) -- (f13);
\path[draw, line width=0.5pt] (v5) -- (f14);
\path[draw, line width=0.5pt] (v13) -- (f14);
\path[draw, line width=0.5pt] (v12) -- (f15);
\path[draw, line width=0.5pt] (v14) -- (f15);
\path[draw, line width=0.5pt] (v12) -- (f16);
\path[draw, line width=0.5pt] (v13) -- (f16);
\path[draw, line width=0.5pt] (v8) -- (f17);
\path[draw, line width=0.5pt] (v14) -- (f17);
\node[scale=1, fill=def_blue!40][circle, inner sep=2.8pt, draw] (v1) at (0, 0) {};
\node[scale=1, fill=def_blue!40][circle, inner sep=2.8pt, draw] (v2) at (1, 0) {};
\node[scale=1, fill=def_blue!40][circle, inner sep=2.8pt, draw] (v3) at (2, 0) {};
\node[scale=1, fill=def_blue!40][circle, inner sep=2.8pt, draw] (v4) at (3, 0) {};
\node[scale=1, fill=def_purple!30][circle, inner sep=2.8pt, draw] (v5) at (4, 0) {};
\node[scale=1, fill=def_purple!30][circle, inner sep=2.8pt, draw] (v6) at (5, 0) {};
\node[scale=1, fill=def_red!30][circle, inner sep=2.8pt, draw] (v7) at (6, 0) {};
\node[scale=1, fill=def_red!30][circle, inner sep=2.8pt, draw] (v8) at (7, 0) {};
\node[scale=1, fill=def_blue!40][circle, inner sep=2.8pt, draw] (v9) at (3, -1) {};
\node[scale=1, fill=def_purple!30][circle, inner sep=2.8pt, draw] (v10) at (4, -1) {};
\node[scale=1, fill=def_purple!30][circle, inner sep=2.8pt, draw] (v11) at (5, -1) {};
\node[scale=1, fill=def_blue!40][circle, inner sep=2.8pt, draw] (v12) at (2, 1) {};
\node[scale=1, fill=def_purple!30][circle, inner sep=2.8pt, draw] (v13) at (4, 1) {};
\node[scale=1, fill=def_blue!40][circle, inner sep=2.8pt, draw] (v14) at (2, 2) {};
\draw [def_blue, fill=def_blue] (f1) circle (0.03);
\draw [def_blue, fill=def_blue] (f2) circle (0.03);
\draw [def_blue, fill=def_blue] (f3) circle (0.03);
\draw [def_blue, fill=def_blue] (f4) circle (0.03);
\draw [def_purple, fill=def_purple] (f5) circle (0.03);
\draw [def_red, fill=def_red] (f6) circle (0.03);
\draw [def_red, fill=def_red] (f7) circle (0.03);
\draw [def_blue, fill=def_blue] (f8) circle (0.03);
\draw [def_purple, fill=def_purple] (f9) circle (0.03);
\draw [def_purple, fill=def_purple] (f10) circle (0.03);
\draw [def_blue, fill=def_blue] (f11) circle (0.03);
\draw [def_purple, fill=def_purple] (f12) circle (0.03);
\draw [def_blue, fill=def_blue] (f13) circle (0.03);
\draw [def_purple, fill=def_purple] (f14) circle (0.03);
\draw [def_blue, fill=def_blue] (f15) circle (0.03);
\draw [def_blue, fill=def_blue] (f16) circle (0.03);
\draw [black, fill=black] (f17) circle (0.03);
\draw [def_blue] (2.5, 0) circle (2.9);
\draw [def_red] (5.4, 0) circle (1.9);
\node[scale=1, draw=none, text=def_red] at (6.0, -0.5) {$\vn$};
\node[scale=1, draw=none, text=def_red] at (6.6, -0.5) {$\zns$};
\node[scale=1, draw=none, text=def_purple] at (4.1, -1.5) {$\vs$};
\node[scale=1, draw=none, text=def_purple] at (4.6, -1.5) {$\zs$};
\node[scale=1, draw=none, text=def_blue] at (1.0, -0.5) {$\vh$};
\node[scale=1, draw=none, text=def_blue] at (1.6, -0.5) {$\zh$};
\node[scale=1, draw=none, ] at (4.6, 1.35) {$\zlc$};
\node[scale=1, draw=none, ] at (2, 2.4) {$\vlc$};
\node[scale=1, draw=none, text=def_blue] at (2.5, -2.2) {server};
\node[scale=1, draw=none, text=def_red] at (5.4, -2.2) {client};
\end{tikzpicture}

%% file: graphs/marginalization.tikz
\begin{tikzpicture}
\definecolor {plum} {rgb} { 0.5,0,0.5 };
\definecolor {pink} {rgb} { 1,0,0.8 };
\definecolor {def_purple} {rgb} { 0.6,0.1,1.0 };
\definecolor {def_blue} {rgb} { 0.12156862745098039,0.4666666666666667,0.7058823529411765 };
\definecolor {def_red} {rgb} { 0.8392156862745098,0.15294117647058825,0.1568627450980392 };
\coordinate (g1v1_coord) at (0, 0);
\coordinate (g1v2_coord) at (1, 0);
\coordinate (g1v3_coord) at (2, 0);
\coordinate (g1v4_coord) at (3, 0);
\coordinate (g1v5_coord) at (4, 0);
\coordinate (g1v6_coord) at (5, 0);
\coordinate (g1v7_coord) at (6, 0);
\coordinate (g1v8_coord) at (7, 0);
\coordinate (g1v9_coord) at (3, -1);
\coordinate (g1v10_coord) at (4, -1);
\coordinate (g1v11_coord) at (5, -1);
\coordinate (g1v12_coord) at (2, 1);
\coordinate (g1v13_coord) at (4, 1);
\coordinate (g1v14_coord) at (2, 2);
\coordinate (g1f1_coord) at (0.5, 0.0);
\coordinate (g1f2_coord) at (1.5, 0.0);
\coordinate (g1f3_coord) at (2.5, 0.0);
\coordinate (g1f4_coord) at (3.5, 0.0);
\coordinate (g1f5_coord) at (4.5, 0.0);
\coordinate (g1f6_coord) at (5.5, 0.0);
\coordinate (g1f7_coord) at (6.5, 0.0);
\coordinate (g1f8_coord) at (3.0, -0.5);
\coordinate (g1f9_coord) at (4.0, -0.5);
\coordinate (g1f10_coord) at (5.0, -0.5);
\coordinate (g1f11_coord) at (3.5, -1.0);
\coordinate (g1f12_coord) at (4.5, -1.0);
\coordinate (g1f13_coord) at (2.0, 0.5);
\coordinate (g1f14_coord) at (4.0, 0.5);
\coordinate (g1f15_coord) at (2.0, 1.5);
\coordinate (g1f16_coord) at (3.0, 1.0);
\path[draw=def_blue, line width=0.3pt] (g1v1_coord) -- (g1f1_coord);
\path[draw=def_blue, line width=0.3pt] (g1v2_coord) -- (g1f1_coord);
\node[circle, scale=0.3, fill=def_blue] at (g1f1_coord) (f1) {};
\path[draw=def_blue, line width=0.3pt] (g1v2_coord) -- (g1f2_coord);
\path[draw=def_blue, line width=0.3pt] (g1v3_coord) -- (g1f2_coord);
\node[circle, scale=0.3, fill=def_blue] at (g1f2_coord) (f2) {};
\path[draw=def_blue, line width=0.3pt] (g1v3_coord) -- (g1f3_coord);
\path[draw=def_blue, line width=0.3pt] (g1v4_coord) -- (g1f3_coord);
\node[circle, scale=0.3, fill=def_blue] at (g1f3_coord) (f3) {};
\path[draw=def_blue, line width=0.3pt] (g1v4_coord) -- (g1f4_coord);
\path[draw=def_blue, line width=0.3pt] (g1v5_coord) -- (g1f4_coord);
\node[circle, scale=0.3, fill=def_blue] at (g1f4_coord) (f4) {};
\path[draw=def_purple, line width=0.3pt] (g1v5_coord) -- (g1f5_coord);
\path[draw=def_purple, line width=0.3pt] (g1v6_coord) -- (g1f5_coord);
\node[circle, scale=0.3, fill=def_purple] at (g1f5_coord) (f5) {};
\path[draw=def_blue, line width=0.3pt] (g1v4_coord) -- (g1f8_coord);
\path[draw=def_blue, line width=0.3pt] (g1v9_coord) -- (g1f8_coord);
\node[circle, scale=0.3, fill=def_blue] at (g1f8_coord) (f8) {};
\path[draw=def_purple, line width=0.3pt] (g1v5_coord) -- (g1f9_coord);
\path[draw=def_purple, line width=0.3pt] (g1v10_coord) -- (g1f9_coord);
\node[circle, scale=0.3, fill=def_purple] at (g1f9_coord) (f9) {};
\path[draw=def_purple, line width=0.3pt] (g1v6_coord) -- (g1f10_coord);
\path[draw=def_purple, line width=0.3pt] (g1v11_coord) -- (g1f10_coord);
\node[circle, scale=0.3, fill=def_purple] at (g1f10_coord) (f10) {};
\path[draw=def_blue, line width=0.3pt] (g1v9_coord) -- (g1f11_coord);
\path[draw=def_blue, line width=0.3pt] (g1v10_coord) -- (g1f11_coord);
\node[circle, scale=0.3, fill=def_blue] at (g1f11_coord) (f11) {};
\path[draw=def_purple, line width=0.3pt] (g1v10_coord) -- (g1f12_coord);
\path[draw=def_purple, line width=0.3pt] (g1v11_coord) -- (g1f12_coord);
\node[circle, scale=0.3, fill=def_purple] at (g1f12_coord) (f12) {};
\path[draw=def_blue, line width=0.3pt] (g1v3_coord) -- (g1f13_coord);
\path[draw=def_blue, line width=0.3pt] (g1v12_coord) -- (g1f13_coord);
\node[circle, scale=0.3, fill=def_blue] at (g1f13_coord) (f13) {};
\path[draw=def_purple, line width=0.3pt] (g1v5_coord) -- (g1f14_coord);
\path[draw=def_purple, line width=0.3pt] (g1v13_coord) -- (g1f14_coord);
\node[circle, scale=0.3, fill=def_purple] at (g1f14_coord) (f14) {};
\path[draw=def_blue, line width=0.3pt] (g1v12_coord) -- (g1f15_coord);
\path[draw=def_blue, line width=0.3pt] (g1v14_coord) -- (g1f15_coord);
\node[circle, scale=0.3, fill=def_blue] at (g1f15_coord) (f15) {};
\path[draw=def_blue, line width=0.3pt] (g1v12_coord) -- (g1f16_coord);
\path[draw=def_blue, line width=0.3pt] (g1v13_coord) -- (g1f16_coord);
\node[circle, scale=0.3, fill=def_blue] at (g1f16_coord) (f16) {};
\node[scale=1, fill=def_blue!40][circle, inner sep=2.8pt, draw, very thin] at (g1v1_coord) (v1) {};
\node[scale=1, fill=def_blue!40][circle, inner sep=2.8pt, draw, very thin] at (g1v2_coord) (v2) {};
\node[scale=1, fill=def_blue!40][circle, inner sep=2.8pt, draw, very thin] at (g1v3_coord) (v3) {};
\node[scale=1, fill=def_blue!40][circle, inner sep=2.8pt, draw, very thin] at (g1v4_coord) (v4) {};
\node[scale=1, fill=def_purple!30][circle, inner sep=2.8pt, draw, very thin] at (g1v5_coord) (v5) {};
\node[scale=1, fill=def_purple!30][circle, inner sep=2.8pt, draw, very thin] at (g1v6_coord) (v6) {};
\node[scale=1, fill=def_blue!40][circle, inner sep=2.8pt, draw, very thin] at (g1v9_coord) (v9) {};
\node[scale=1, fill=def_purple!30][circle, inner sep=2.8pt, draw, very thin] at (g1v10_coord) (v10) {};
\node[scale=1, fill=def_purple!30][circle, inner sep=2.8pt, draw, very thin] at (g1v11_coord) (v11) {};
\node[scale=1, fill=def_blue!40][circle, inner sep=2.8pt, draw, very thin] at (g1v12_coord) (v12) {};
\node[scale=1, fill=def_purple!30][circle, inner sep=2.8pt, draw, very thin] at (g1v13_coord) (v13) {};
\node[scale=1, fill=def_blue!40][circle, inner sep=2.8pt, draw, very thin] at (g1v14_coord) (v14) {};
\coordinate (g2v1_coord) at (5, 0);
\coordinate (g2v2_coord) at (6, 0);
\coordinate (g2v3_coord) at (7, 0);
\coordinate (g2v4_coord) at (8, 0);
\coordinate (g2v5_coord) at (9, 0);
\coordinate (g2v6_coord) at (10, 0);
\coordinate (g2v7_coord) at (11, 0);
\coordinate (g2v8_coord) at (12, 0);
\coordinate (g2v9_coord) at (8, -1);
\coordinate (g2v10_coord) at (9, -1);
\coordinate (g2v11_coord) at (10, -1);
\coordinate (g2v12_coord) at (7, 1);
\coordinate (g2v13_coord) at (9, 1);
\coordinate (g2v14_coord) at (7, 2);
\coordinate (g2f1_coord) at (5.5, 0.0);
\coordinate (g2f2_coord) at (6.5, 0.0);
\coordinate (g2f3_coord) at (7.5, 0.0);
\coordinate (g2f4_coord) at (8.5, 0.0);
\coordinate (g2f5_coord) at (9.5, 0.0);
\coordinate (g2f6_coord) at (10.5, 0.0);
\coordinate (g2f7_coord) at (11.5, 0.0);
\coordinate (g2f8_coord) at (8.0, -0.5);
\coordinate (g2f9_coord) at (9.0, -0.5);
\coordinate (g2f10_coord) at (10.0, -0.5);
\coordinate (g2f11_coord) at (8.5, -1.0);
\coordinate (g2f12_coord) at (9.5, -1.0);
\coordinate (g2f13_coord) at (7.0, 0.5);
\coordinate (g2f14_coord) at (9.0, 0.5);
\coordinate (g2f15_coord) at (7.0, 1.5);
\coordinate (g2f16_coord) at (8.0, 1.0);
\coordinate (g2margold_coord) at (7, 0);
\path[draw=def_blue, line width=0.3pt] (g2v4_coord) -- (g2f4_coord);
\path[draw=def_blue, line width=0.3pt] (g2v5_coord) -- (g2f4_coord);
\node[circle, scale=0.3, fill=def_blue] at (g2f4_coord) (f4) {};
\path[draw=def_purple, line width=0.3pt] (g2v5_coord) -- (g2f5_coord);
\path[draw=def_purple, line width=0.3pt] (g2v6_coord) -- (g2f5_coord);
\node[circle, scale=0.3, fill=def_purple] at (g2f5_coord) (f5) {};
\path[draw=def_red, line width=0.3pt] (g2v6_coord) -- (g2f6_coord);
\path[draw=def_red, line width=0.3pt] (g2v7_coord) -- (g2f6_coord);
\node[circle, scale=0.3, fill=def_red] at (g2f6_coord) (f6) {};
\path[draw=def_red, line width=0.3pt] (g2v7_coord) -- (g2f7_coord);
\path[draw=def_red, line width=0.3pt] (g2v8_coord) -- (g2f7_coord);
\node[circle, scale=0.3, fill=def_red] at (g2f7_coord) (f7) {};
\path[draw=def_blue, line width=0.3pt] (g2v4_coord) -- (g2f8_coord);
\path[draw=def_blue, line width=0.3pt] (g2v9_coord) -- (g2f8_coord);
\node[circle, scale=0.3, fill=def_blue] at (g2f8_coord) (f8) {};
\path[draw=def_purple, line width=0.3pt] (g2v5_coord) -- (g2f9_coord);
\path[draw=def_purple, line width=0.3pt] (g2v10_coord) -- (g2f9_coord);
\node[circle, scale=0.3, fill=def_purple] at (g2f9_coord) (f9) {};
\path[draw=def_purple, line width=0.3pt] (g2v6_coord) -- (g2f10_coord);
\path[draw=def_purple, line width=0.3pt] (g2v11_coord) -- (g2f10_coord);
\node[circle, scale=0.3, fill=def_purple] at (g2f10_coord) (f10) {};
\path[draw=def_blue, line width=0.3pt] (g2v9_coord) -- (g2f11_coord);
\path[draw=def_blue, line width=0.3pt] (g2v10_coord) -- (g2f11_coord);
\node[circle, scale=0.3, fill=def_blue] at (g2f11_coord) (f11) {};
\path[draw=def_purple, line width=0.3pt] (g2v10_coord) -- (g2f12_coord);
\path[draw=def_purple, line width=0.3pt] (g2v11_coord) -- (g2f12_coord);
\node[circle, scale=0.3, fill=def_purple] at (g2f12_coord) (f12) {};
\path[draw=def_purple, line width=0.3pt] (g2v5_coord) -- (g2f14_coord);
\path[draw=def_purple, line width=0.3pt] (g2v13_coord) -- (g2f14_coord);
\node[circle, scale=0.3, fill=def_purple] at (g2f14_coord) (f14) {};
\path[draw=def_blue, line width=0.3pt] (g2v4_coord) -- (g2margold_coord);
\path[draw=def_blue, line width=0.3pt] (g2v9_coord) -- (g2margold_coord);
\path[draw=def_blue, line width=0.3pt] (g2v13_coord) -- (g2margold_coord);
\node[circle, scale=0.3, fill=def_blue] at (g2margold_coord) (margold) {};
\node[scale=1, fill=def_blue!40][circle, inner sep=2.8pt, draw, very thin] at (g2v4_coord) (v4) {};
\node[scale=1, fill=def_purple!30][circle, inner sep=2.8pt, draw, very thin] at (g2v5_coord) (v5) {};
\node[scale=1, fill=def_purple!30][circle, inner sep=2.8pt, draw, very thin] at (g2v6_coord) (v6) {};
\node[scale=1, fill=def_red!30][circle, inner sep=2.8pt, draw, very thin] at (g2v7_coord) (v7) {};
\node[scale=1, fill=def_red!30][circle, inner sep=2.8pt, draw, very thin] at (g2v8_coord) (v8) {};
\node[scale=1, fill=def_blue!40][circle, inner sep=2.8pt, draw, very thin] at (g2v9_coord) (v9) {};
\node[scale=1, fill=def_purple!30][circle, inner sep=2.8pt, draw, very thin] at (g2v10_coord) (v10) {};
\node[scale=1, fill=def_purple!30][circle, inner sep=2.8pt, draw, very thin] at (g2v11_coord) (v11) {};
\node[scale=1, fill=def_purple!30][circle, inner sep=2.8pt, draw, very thin] at (g2v13_coord) (v13) {};
\coordinate (g4v1_coord) at (5, -5);
\coordinate (g4v2_coord) at (6, -5);
\coordinate (g4v3_coord) at (7, -5);
\coordinate (g4v4_coord) at (8, -5);
\coordinate (g4v5_coord) at (9, -5);
\coordinate (g4v6_coord) at (10, -5);
\coordinate (g4v7_coord) at (11, -5);
\coordinate (g4v8_coord) at (12, -5);
\coordinate (g4v9_coord) at (8, -6);
\coordinate (g4v10_coord) at (9, -6);
\coordinate (g4v11_coord) at (10, -6);
\coordinate (g4v12_coord) at (7, -4);
\coordinate (g4v13_coord) at (9, -4);
\coordinate (g4v14_coord) at (7, -3);
\coordinate (g4f1_coord) at (5.5, -5.0);
\coordinate (g4f2_coord) at (6.5, -5.0);
\coordinate (g4f3_coord) at (7.5, -5.0);
\coordinate (g4f4_coord) at (8.5, -5.0);
\coordinate (g4f5_coord) at (9.5, -5.0);
\coordinate (g4f6_coord) at (10.5, -5.0);
\coordinate (g4f7_coord) at (11.5, -5.0);
\coordinate (g4f8_coord) at (8.0, -5.5);
\coordinate (g4f9_coord) at (9.0, -5.5);
\coordinate (g4f10_coord) at (10.0, -5.5);
\coordinate (g4f11_coord) at (8.5, -6.0);
\coordinate (g4f12_coord) at (9.5, -6.0);
\coordinate (g4f13_coord) at (7.0, -4.5);
\coordinate (g4f14_coord) at (9.0, -4.5);
\coordinate (g4f15_coord) at (7.0, -3.5);
\coordinate (g4f16_coord) at (8.0, -4.0);
\coordinate (g4margnew_coord) at (8, -5);
\path[draw=def_purple, line width=0.3pt] (g4v5_coord) -- (g4f5_coord);
\path[draw=def_purple, line width=0.3pt] (g4v6_coord) -- (g4f5_coord);
\node[circle, scale=0.3, fill=def_purple] at (g4f5_coord) (f5) {};
\path[draw=def_red, line width=0.3pt] (g4v6_coord) -- (g4f6_coord);
\path[draw=def_red, line width=0.3pt] (g4v7_coord) -- (g4f6_coord);
\node[circle, scale=0.3, fill=def_red] at (g4f6_coord) (f6) {};
\path[draw=def_red, line width=0.3pt] (g4v7_coord) -- (g4f7_coord);
\path[draw=def_red, line width=0.3pt] (g4v8_coord) -- (g4f7_coord);
\node[circle, scale=0.3, fill=def_red] at (g4f7_coord) (f7) {};
\path[draw=def_purple, line width=0.3pt] (g4v5_coord) -- (g4f9_coord);
\path[draw=def_purple, line width=0.3pt] (g4v10_coord) -- (g4f9_coord);
\node[circle, scale=0.3, fill=def_purple] at (g4f9_coord) (f9) {};
\path[draw=def_purple, line width=0.3pt] (g4v6_coord) -- (g4f10_coord);
\path[draw=def_purple, line width=0.3pt] (g4v11_coord) -- (g4f10_coord);
\node[circle, scale=0.3, fill=def_purple] at (g4f10_coord) (f10) {};
\path[draw=def_purple, line width=0.3pt] (g4v10_coord) -- (g4f12_coord);
\path[draw=def_purple, line width=0.3pt] (g4v11_coord) -- (g4f12_coord);
\node[circle, scale=0.3, fill=def_purple] at (g4f12_coord) (f12) {};
\path[draw=def_purple, line width=0.3pt] (g4v5_coord) -- (g4f14_coord);
\path[draw=def_purple, line width=0.3pt] (g4v13_coord) -- (g4f14_coord);
\node[circle, scale=0.3, fill=def_purple] at (g4f14_coord) (f14) {};
\path[draw=pink, line width=0.3pt] (g4v5_coord) -- (g4margnew_coord);
\path[draw=pink, line width=0.3pt] (g4v10_coord) -- (g4margnew_coord);
\path[draw=pink, line width=0.3pt] (g4v13_coord) -- (g4margnew_coord);
\node[circle, scale=0.3, fill=pink] at (g4margnew_coord) (margnew) {};
\node[scale=1, fill=def_purple!30][circle, inner sep=2.8pt, draw, very thin] at (g4v5_coord) (v5) {};
\node[scale=1, fill=def_purple!30][circle, inner sep=2.8pt, draw, very thin] at (g4v6_coord) (v6) {};
\node[scale=1, fill=def_red!30][circle, inner sep=2.8pt, draw, very thin] at (g4v7_coord) (v7) {};
\node[scale=1, fill=def_red!30][circle, inner sep=2.8pt, draw, very thin] at (g4v8_coord) (v8) {};
\node[scale=1, fill=def_purple!30][circle, inner sep=2.8pt, draw, very thin] at (g4v10_coord) (v10) {};
\node[scale=1, fill=def_purple!30][circle, inner sep=2.8pt, draw, very thin] at (g4v11_coord) (v11) {};
\node[scale=1, fill=def_purple!30][circle, inner sep=2.8pt, draw, very thin] at (g4v13_coord) (v13) {};
\coordinate (g3v1_coord) at (0, -5);
\coordinate (g3v2_coord) at (1, -5);
\coordinate (g3v3_coord) at (2, -5);
\coordinate (g3v4_coord) at (3, -5);
\coordinate (g3v5_coord) at (4, -5);
\coordinate (g3v6_coord) at (5, -5);
\coordinate (g3v7_coord) at (6, -5);
\coordinate (g3v8_coord) at (7, -5);
\coordinate (g3v9_coord) at (3, -6);
\coordinate (g3v10_coord) at (4, -6);
\coordinate (g3v11_coord) at (5, -6);
\coordinate (g3v12_coord) at (2, -4);
\coordinate (g3v13_coord) at (4, -4);
\coordinate (g3v14_coord) at (2, -3);
\coordinate (g3f1_coord) at (0.5, -5.0);
\coordinate (g3f2_coord) at (1.5, -5.0);
\coordinate (g3f3_coord) at (2.5, -5.0);
\coordinate (g3f4_coord) at (3.5, -5.0);
\coordinate (g3f5_coord) at (4.5, -5.0);
\coordinate (g3f6_coord) at (5.5, -5.0);
\coordinate (g3f7_coord) at (6.5, -5.0);
\coordinate (g3f8_coord) at (3.0, -5.5);
\coordinate (g3f10_coord) at (5.0, -5.5);
\coordinate (g3f11_coord) at (3.5, -6.0);
\coordinate (g3f12_coord) at (4.5, -6.0);
\coordinate (g3f13_coord) at (2.0, -4.5);
\coordinate (g3f15_coord) at (2.0, -3.5);
\coordinate (g3f16_coord) at (3.0, -4.0);
\coordinate (g3margnew_coord) at (3, -5);
\path[draw=pink, line width=0.3pt] (g3v5_coord) -- (g3margnew_coord);
\path[draw=pink, line width=0.3pt] (g3v10_coord) -- (g3margnew_coord);
\path[draw=pink, line width=0.3pt] (g3v13_coord) -- (g3margnew_coord);
\node[circle, scale=0.3, fill=pink] at (g3margnew_coord) (margnew) {};
\node[scale=1, fill=def_purple!30][circle, inner sep=2.8pt, draw, very thin] at (g3v5_coord) (v5) {};
\node[scale=1, fill=def_purple!30][circle, inner sep=2.8pt, draw, very thin] at (g3v10_coord) (v10) {};
\node[scale=1, fill=def_purple!30][circle, inner sep=2.8pt, draw, very thin] at (g3v13_coord) (v13) {};
\path[->, >=stealth, draw=black, line width=0.3pt] (4, -2) -- (4, -3);
\path[->, >=stealth, draw=black, line width=0.3pt] (6, -5) -- (7, -5);
\path[->, >=stealth, draw=black, line width=0.3pt] (9, -2) -- (9, -3);
\node[scale=1, fill=white][circle, inner sep=0.8pt, draw=black] at (3.5, -2.5) {1};
\node[scale=1, fill=white][circle, inner sep=0.8pt, draw=black] at (6.5, -4.5) {2};
\node[scale=1, fill=white][circle, inner sep=0.8pt, draw=black] at (8.5, -2.5) {3};
\end{tikzpicture}

%% file: graphs/communication.tikz
\begin{tikzpicture}
\path[->, >=stealth, draw=black, line width=0.3pt] (-0.2, -2) -- (6, -2);
\path[draw=black, line width=0.3pt] (0, -2) -- (0, -1.9);
\node[scale=0.5, draw=none] at (0, -2.1) {$t_1$};
\path[->, >=stealth, draw=black, line width=0.3pt] (0, 0) -- (0, -0.3);
\path[->, >=stealth, draw=black, line width=0.3pt] (0, -0.5) -- (0.3, -0.5);
\path[dashed, ->, >=stealth, draw=gray, line width=0.3pt] (0.3, -0.5) -- (0.3, -0.7);
\path[->, >=stealth, draw=black, line width=0.3pt] (0.3, -0.7) -- (0.6, -0.7);
\path[->, >=stealth, draw=black, line width=0.3pt] (0.2, -1.3) -- (0.5, -1.3);
\path[dashed, ->, >=stealth, draw=gray, line width=0.3pt] (0.5, -1.3) -- (0.5, -1.5);
\path[->, >=stealth, draw=black, line width=0.3pt] (0, -0.5) -- (0.2, -1.3);
\path[draw=black, line width=0.3pt] (1, -2) -- (1, -1.9);
\node[scale=0.5, draw=none] at (1, -2.1) {$t_2$};
\path[->, >=stealth, draw=blue, line width=0.3pt] (1, 0) -- (1, -0.3);
\path[->, >=stealth, draw=blue, line width=0.3pt] (1, -0.5) -- (1.3, -0.5);
\path[dashed, ->, >=stealth, draw=gray, line width=0.3pt] (1.3, -0.5) -- (1.3, -0.7);
\path[->, >=stealth, draw=blue, line width=0.3pt] (1.3, -0.7) -- (1.6, -0.7);
\path[->, >=stealth, draw=blue, line width=0.3pt] (1.2, -1.3) -- (1.5, -1.3);
\path[dashed, ->, >=stealth, draw=gray, line width=0.3pt] (1.5, -1.3) -- (1.5, -1.5);
\path[->, >=stealth, draw=blue, line width=0.3pt] (1, -0.5) -- (1.2, -1.3);
\path[draw=black, line width=0.3pt] (2, -2) -- (2, -1.9);
\node[scale=0.5, draw=none] at (2, -2.1) {$t_3$};
\path[->, >=stealth, draw=blue, line width=0.3pt] (2, 0) -- (2, -0.3);
\path[->, >=stealth, draw=blue, line width=0.3pt] (2, -0.5) -- (2.3, -0.5);
\path[dashed, ->, >=stealth, draw=gray, line width=0.3pt] (2.3, -0.5) -- (2.3, -0.7);
\path[->, >=stealth, draw=blue, line width=0.3pt] (2.3, -0.7) -- (2.6, -0.7);
\path[->, >=stealth, draw=blue, line width=0.3pt] (2.2, -1.3) -- (2.5, -1.3);
\path[dashed, ->, >=stealth, draw=gray, line width=0.3pt] (2.5, -1.3) -- (2.5, -1.5);
\path[->, >=stealth, draw=blue, line width=0.3pt] (2, -0.5) -- (2.2, -1.3);
\path[draw=black, line width=0.3pt] (3, -2) -- (3, -1.9);
\node[scale=0.5, draw=none] at (3, -2.1) {$t_4$};
\path[->, >=stealth, draw=black, line width=0.3pt] (3, 0) -- (3, -0.3);
\path[->, >=stealth, draw=black, line width=0.3pt] (3, -0.5) -- (3.3, -0.5);
\path[dashed, ->, >=stealth, draw=gray, line width=0.3pt] (3.3, -0.5) -- (3.3, -0.7);
\path[->, >=stealth, draw=black, line width=0.3pt] (3.3, -0.7) -- (3.6, -0.7);
\path[->, >=stealth, draw=black, line width=0.3pt] (3.2, -1.3) -- (3.5, -1.3);
\path[dashed, ->, >=stealth, draw=gray, line width=0.3pt] (3.5, -1.3) -- (3.5, -1.5);
\path[->, >=stealth, draw=black, line width=0.3pt] (3, -0.5) -- (3.2, -1.3);
\path[draw=black, line width=0.3pt] (4, -2) -- (4, -1.9);
\node[scale=0.5, draw=none] at (4, -2.1) {$t_5$};
\path[->, >=stealth, draw=black, line width=0.3pt] (4, 0) -- (4, -0.3);
\path[->, >=stealth, draw=black, line width=0.3pt] (4, -0.5) -- (4.3, -0.5);
\path[dashed, ->, >=stealth, draw=gray, line width=0.3pt] (4.3, -0.5) -- (4.3, -0.7);
\path[->, >=stealth, draw=black, line width=0.3pt] (4.3, -0.7) -- (4.6, -0.7);
\path[->, >=stealth, draw=black, line width=0.3pt] (4.2, -1.3) -- (4.5, -1.3);
\path[dashed, ->, >=stealth, draw=gray, line width=0.3pt] (4.5, -1.3) -- (4.5, -1.5);
\path[->, >=stealth, draw=black, line width=0.3pt] (4, -0.5) -- (4.2, -1.3);
\path[draw=black, line width=0.3pt] (5, -2) -- (5, -1.9);
\node[scale=0.5, draw=none] at (5, -2.1) {$t_6$};
\path[->, >=stealth, draw=black, line width=0.3pt] (5, 0) -- (5, -0.3);
\path[->, >=stealth, draw=black, line width=0.3pt] (5, -0.5) -- (5.3, -0.5);
\path[dashed, ->, >=stealth, draw=gray, line width=0.3pt] (5.3, -0.5) -- (5.3, -0.7);
\path[->, >=stealth, draw=black, line width=0.3pt] (5.3, -0.7) -- (5.6, -0.7);
\path[->, >=stealth, draw=black, line width=0.3pt] (5.2, -1.3) -- (5.5, -1.3);
\path[dashed, ->, >=stealth, draw=gray, line width=0.3pt] (5.5, -1.3) -- (5.5, -1.5);
\path[->, >=stealth, draw=black, line width=0.3pt] (5, -0.5) -- (5.2, -1.3);
\path[->, >=stealth, draw=black, line width=0.3pt] (0.5, -1.5) -- (2.5, -1.5);
\path[->, >=stealth, draw=blue, line width=0.3pt] (2.5, -1.5) -- (4.5, -1.5);
\path[->, >=stealth, draw=black, line width=0.3pt] (2.5, -1.5) -- (2.8, -0.7);
\path[->, >=stealth, draw=blue, line width=0.3pt] (4.5, -1.5) -- (4.8, -0.7);
\path[dashed, ->, >=stealth, draw=red, line width=0.3pt] (1.5, -1.3) -- (1.7, -0.7);
\node[scale=0.5, fill=white][circle, inner sep=0.8pt, draw=black] at (0.8, -0.2) {1};
\node[scale=0.5, fill=white][circle, inner sep=0.8pt, draw=black] at (1.1, -0.35) {2};
\node[scale=0.5, fill=white][circle, inner sep=0.8pt, draw=black] at (1.45, -0.55) {3};
\node[scale=0.5, fill=white][circle, inner sep=0.8pt, draw=black] at (0.95, -0.9) {4};
\node[scale=0.5, fill=white][circle, inner sep=0.8pt, draw=black] at (1.35, -1.15) {5};
\node[scale=0.5, fill=white][circle, inner sep=0.8pt, draw=black] at (3.5, -1.65) {6};
\node[scale=0.5, fill=white][circle, inner sep=0.8pt, draw=black] at (4.75, -1.25) {7};
\node[scale=0.5, fill=white][circle, inner sep=0.8pt, draw=black] at (1.7, -1.1) {8};
\node[scale=0.7, draw=none] at (-1.1, -0.45) {device front-end};
\node[scale=0.7, draw=none] at (-1.1, -0.75) {device back-end};
\node[scale=0.7, draw=none] at (-1.1, -1.25) {server front-end};
\node[scale=0.7, draw=none] at (-1.1, -1.55) {server back-end};
\node[scale=0.7, draw=none] at (5.8, -2.2) {time};
\end{tikzpicture}

%% file: graphs/spatial_fg.tikz
\begin{tikzpicture}
\definecolor {plum} {rgb} { 0.5,0,0.5 };
\definecolor {pink} {rgb} { 1,0,0.8 };
\definecolor {def_purple} {rgb} { 0.6,0.1,1.0 };
\definecolor {def_blue} {rgb} { 0.12156862745098039,0.4666666666666667,0.7058823529411765 };
\definecolor {def_red} {rgb} { 0.8392156862745098,0.15294117647058825,0.1568627450980392 };
\coordinate (v1) at (0, 0);
\coordinate (v2) at (1, 0);
\coordinate (v3) at (2, 0);
\coordinate (v4) at (3, 0);
\coordinate (v5) at (4, 0);
\coordinate (v6) at (2, 1);
\coordinate (v7) at (3, 1);
\coordinate (v8) at (4, 1);
\coordinate (v9) at (2, -1);
\coordinate (v10) at (4, -1);
\coordinate (v11) at (1, 2);
\coordinate (v12) at (5, 1);
\coordinate (v13) at (5, -1);
\coordinate (v14) at (2, 2.5);
\coordinate (v15) at (4.5, 2.5);
\coordinate (f1) at (0.5, 0.0);
\coordinate (f2) at (1.5, 0.0);
\coordinate (f3) at (2.5, 0.0);
\coordinate (f4) at (3.5, 0.0);
\coordinate (f5) at (2.0, 0.5);
\coordinate (f6) at (3.0, 0.5);
\coordinate (f7) at (4.0, 0.5);
\coordinate (f8) at (2.5, 1.0);
\coordinate (f9) at (3.5, 1.0);
\coordinate (f10) at (1.5, -0.5);
\coordinate (f11) at (2.0, -0.5);
\coordinate (f12) at (2.5, -0.5);
\coordinate (f13) at (3.0, -1.0);
\coordinate (f14) at (4.0, -0.5);
\coordinate (f15) at (0.5, 1.0);
\coordinate (f16) at (1.5, 1.5);
\coordinate (f17) at (4.5, 1.0);
\coordinate (f18) at (4.5, 0.5);
\coordinate (f19) at (4.5, -1.0);
\coordinate (f20) at (1.5, 2.25);
\coordinate (f21) at (3.25, 2.5);
\coordinate (f22) at (4.75, 1.75);
\coordinate (f23) at (5.0, 0.0);
\path[draw, line width=0.5pt] (v1) -- (f1);
\path[draw, line width=0.5pt] (v2) -- (f1);
\path[draw, line width=0.5pt] (v2) -- (f2);
\path[draw, line width=0.5pt] (v3) -- (f2);
\path[draw, line width=0.5pt] (v3) -- (f3);
\path[draw, line width=0.5pt] (v4) -- (f3);
\path[draw, line width=0.5pt] (v4) -- (f4);
\path[draw, line width=0.5pt] (v5) -- (f4);
\path[draw, line width=0.5pt] (v3) -- (f5);
\path[draw, line width=0.5pt] (v6) -- (f5);
\path[draw, line width=0.5pt] (v4) -- (f6);
\path[draw, line width=0.5pt] (v7) -- (f6);
\path[draw, line width=0.5pt] (v5) -- (f7);
\path[draw, line width=0.5pt] (v8) -- (f7);
\path[draw, line width=0.5pt] (v6) -- (f8);
\path[draw, line width=0.5pt] (v7) -- (f8);
\path[draw, line width=0.5pt] (v7) -- (f9);
\path[draw, line width=0.5pt] (v8) -- (f9);
\path[draw, line width=0.5pt] (v2) -- (f10);
\path[draw, line width=0.5pt] (v9) -- (f10);
\path[draw, line width=0.5pt] (v3) -- (f11);
\path[draw, line width=0.5pt] (v9) -- (f11);
\path[draw, line width=0.5pt] (v4) -- (f12);
\path[draw, line width=0.5pt] (v9) -- (f12);
\path[draw, line width=0.5pt] (v9) -- (f13);
\path[draw, line width=0.5pt] (v10) -- (f13);
\path[draw, line width=0.5pt] (v5) -- (f14);
\path[draw, line width=0.5pt] (v10) -- (f14);
\path[draw, line width=0.5pt] (v1) -- (f15);
\path[draw, line width=0.5pt] (v11) -- (f15);
\path[draw, line width=0.5pt] (v6) -- (f16);
\path[draw, line width=0.5pt] (v11) -- (f16);
\path[draw, line width=0.5pt] (v8) -- (f17);
\path[draw, line width=0.5pt] (v12) -- (f17);
\path[draw, line width=0.5pt] (v5) -- (f18);
\path[draw, line width=0.5pt] (v12) -- (f18);
\path[draw, line width=0.5pt] (v10) -- (f19);
\path[draw, line width=0.5pt] (v13) -- (f19);
\path[draw, line width=0.5pt] (v11) -- (f20);
\path[draw, line width=0.5pt] (v14) -- (f20);
\path[draw, line width=0.5pt] (v14) -- (f21);
\path[draw, line width=0.5pt] (v15) -- (f21);
\path[draw, line width=0.5pt] (v15) -- (f22);
\path[draw, line width=0.5pt] (v12) -- (f22);
\path[draw, line width=0.5pt] (v12) -- (f23);
\path[draw, line width=0.5pt] (v13) -- (f23);
\node[scale=1, fill=def_blue!20][circle, inner sep=2.8pt, draw] (v1) at (0, 0) {$x_1$};
\node[scale=1, fill=def_blue!20][circle, inner sep=2.8pt, draw] (v2) at (1, 0) {$x_2$};
\node[scale=1, fill=def_purple!20][circle, inner sep=2.8pt, draw] (v3) at (2, 0) {$x_3$};
\node[scale=1, fill=def_red!40][circle, inner sep=2.8pt, draw] (v4) at (3, 0) {$x_4$};
\node[scale=1, fill=orange!30][circle, inner sep=2.8pt, draw] (v5) at (4, 0) {$x_5$};
\node[scale=1, fill=def_purple!20][circle, inner sep=2.8pt, draw] (v6) at (2, 1) {};
\node[scale=1, fill=def_purple!20][circle, inner sep=2.8pt, draw] (v7) at (3, 1) {};
\node[scale=1, fill=def_purple!20][circle, inner sep=2.8pt, draw] (v8) at (4, 1) {};
\node[scale=1, fill=def_purple!20][circle, inner sep=2.8pt, draw] (v9) at (2, -1) {};
\node[scale=1, fill=def_purple!20][circle, inner sep=2.8pt, draw] (v10) at (4, -1) {};
\node[scale=1, fill=def_blue!20][circle, inner sep=2.8pt, draw] (v11) at (1, 2) {};
\node[scale=1, fill=orange!20][circle, inner sep=2.8pt, draw] (v12) at (5, 1) {};
\node[scale=1, fill=orange!20][circle, inner sep=2.8pt, draw] (v13) at (5, -1) {};
\node[scale=1, fill=def_blue!20][circle, inner sep=2.8pt, draw] (v14) at (2, 2.5) {};
\node[scale=1, fill=def_blue!20][circle, inner sep=2.8pt, draw] (v15) at (4.5, 2.5) {};
\draw [def_blue, fill=def_blue] (f1) circle (0.03);
\draw [def_blue, fill=def_blue] (f2) circle (0.03);
\draw [def_red, fill=def_red] (f3) circle (0.03);
\draw [orange, fill=orange] (f4) circle (0.03);
\draw [def_purple, fill=def_purple] (f5) circle (0.03);
\draw [def_red, fill=def_red] (f6) circle (0.03);
\draw [orange, fill=orange] (f7) circle (0.03);
\draw [def_purple, fill=def_purple] (f8) circle (0.03);
\draw [def_purple, fill=def_purple] (f9) circle (0.03);
\draw [def_blue, fill=def_blue] (f10) circle (0.03);
\draw [def_purple, fill=def_purple] (f11) circle (0.03);
\draw [def_red, fill=def_red] (f12) circle (0.03);
\draw [def_purple, fill=def_purple] (f13) circle (0.03);
\draw [orange, fill=orange] (f14) circle (0.03);
\draw [def_blue, fill=def_blue] (f15) circle (0.03);
\draw [def_blue, fill=def_blue] (f16) circle (0.03);
\draw [orange, fill=orange] (f17) circle (0.03);
\draw [orange, fill=orange] (f18) circle (0.03);
\draw [orange, fill=orange] (f19) circle (0.03);
\draw [def_blue, fill=def_blue] (f20) circle (0.03);
\draw [def_blue, fill=def_blue] (f21) circle (0.03);
\draw [def_blue, fill=def_blue] (f22) circle (0.03);
\draw [orange, fill=orange] (f23) circle (0.03);
\draw [def_red] (3, 0) circle (1.8);
\draw [orange] (4, 0) circle (1.8);
\end{tikzpicture}